\documentclass[11pt]{article}

\usepackage{acl}

\usepackage{times}
\usepackage{tcolorbox}
\usepackage{pifont}

\usepackage[utf8]{inputenc} 
\usepackage[T1]{fontenc}   
\usepackage{hyperref}       
\usepackage{url}
\usepackage{multirow}       
\usepackage{booktabs}  
\usepackage{amssymb}
\usepackage{amsthm}

\usepackage{amsfonts}       
\usepackage{nicefrac}       
\usepackage{tabularx}
\usepackage{longtable}
\usepackage{supertabular}
\usepackage{makecell}
\usepackage{microtype}      
\usepackage{xcolor}         
\usepackage{tabularx}
\usepackage{graphicx}
\graphicspath{ {imgs/} }
\usepackage{svg}
\usepackage{enumitem}
\usepackage{subcaption}
\usepackage{amsthm, amsmath}
\usepackage[linesnumbered,ruled]{algorithm2e}
\usepackage{colortbl}

\definecolor{blue}{RGB}{17,220,247}
\definecolor{purple}{RGB}{163,115,250}
\definecolor{caribbeangreen}{rgb}{0.0, 0.8, 0.6}

\definecolor{GREEN}{RGB}{84,130,53}

\newcommand{\colorg}{\cellcolor{gray!15}}

\usepackage{wrapfig}

\usepackage{todonotes}

\usepackage{bbm}

\usepackage{algorithm}
\usepackage{algpseudocode}
\usepackage[ruled]{algorithm2e}

\newcommand{\name}{\textsc{VerifierFC}}

\definecolor{GREEN}{RGB}{84,130,53}

\usepackage{bbm}

\usepackage{pgfplots}
\pgfplotsset{compat=1.15}
\usepgfplotslibrary{
  statistics,
  colorbrewer,
  groupplots,
}
\usetikzlibrary{
  patterns,
  shapes.geometric,
  decorations.text,
  matrix,
  fit,
  backgrounds,
  positioning,
}
\tikzset{
  fignode/.style={
    outer sep=0.25em,
  }
}
\tikzset{
  framedfignode/.style={
    outer sep=0.25em,
    inner sep=0.5em,
    rounded corners,
    draw,
  }
}
\colorlet{plotColorNeutral}{gray}
\definecolor{plotColor1}{HTML}{f61a1c}
\definecolor{plotColor2}{HTML}{377eb8}
\definecolor{plotColor3}{HTML}{4daf4a}
\definecolor{plotColor4}{HTML}{984ea3}
\definecolor{plotColor5}{HTML}{FFFFCB}
\definecolor{plotColor6}{HTML}{1e90ff}
\colorlet{plotColorNeutral*}{plotColorNeutral!40}
\colorlet{plotColor1*}{plotColor1!60}
\colorlet{plotColor2*}{plotColor2!60}
\colorlet{plotColor3*}{plotColor3!60}
\colorlet{plotColor4*}{plotColor4!60}
\colorlet{plotColor5*}{plotColor5!60}
\colorlet{plotColor6*}{plotColor6!60}
\pgfplotsset{
    colormap={greenred}{HTML=(4daf4a) HTML=(e41a1c)},
    colormap={redgreen}{HTML=(e41a1c) HTML=(4daf4a)}
}

\newcommand{\analysisissue}{reasoning drift}
\newcommand{\Analysisissue}{Reasoning drift}

\newcommand{\cV}{\mathcal{V}}

\theoremstyle{definition}

\title{Think Right, Not More: Test-Time Scaling for Numerical Claim Verification}

\author{ 
 Primakov  \\ \textbf{Chungkham$^*$}  \\ TU Delft \\ \texttt{primakovch} \\ \texttt{@gmail.com}  
\And Venktesh V\Thanks{ Both authors \textbf{contributed equally} to this work} \\ Stockholm University \\ \texttt{venktesh.viswanathan} \\\texttt{@dsv.su.se }
\And Vinay Setty \\ University of Stavanger \\ \texttt{vsetty}\\
\texttt{@acm.org}
\And  Avishek Anand \\ TU Delft \\ \texttt{avishek.anand}\\ \texttt{@tudelft.nl}  }

\begin{document}

\maketitle
\begin{abstract}

Fact-checking real-world claims, particularly numerical claims, is inherently complex that require multistep reasoning and numerical reasoning for verifying diverse aspects of the claim. Although large language models (LLMs) including reasoning models have made tremendous advances, they still fall short on fact-checking real-world claims that require a combination of compositional and numerical reasoning. They are unable to understand nuance of numerical aspects, and are also susceptible to the \textit{\analysisissue} issue, where the model is unable to contextualize diverse information resulting in misinterpretation and backtracking of reasoning process. In this work, we systematically explore scaling test-time compute (TTS) for LLMs on the task of fact-checking complex numerical claims, which entails eliciting multiple reasoning paths from an LLM. We train a verifier model (\name{}) to navigate this space of possible reasoning paths and select one that could lead to the correct verdict. We observe that TTS helps mitigate the \textit{\analysisissue} issue, leading to significant performance gains for fact-checking numerical claims. To improve compute efficiency in TTS, we introduce an adaptive mechanism that performs TTS selectively based on the perceived complexity of the claim. This approach achieves \textbf{1.8$\times$} higher efficiency than standard TTS, while delivering a notable \textbf{18.8\%} performance improvement over single-shot claim verification methods. Our code and data can be found at \url{https://github.com/VenkteshV/VerifierFC}




\end{abstract}

\begin{table}[htb!!] 
\centering
\small
\begin{tcolorbox}[title=Mitigating \textit{\Analysisissue} for Numerical FC]
\small
\medskip\noindent \paragraph{\textbf{Claim}:}\texttt{Diaspora remittances “now stand at an all-time high of over KSh400 billion in 2021”.}

[\textbf{Evidence 1}]:  diaspora remittances stand at an all-time-high at over ksh. \textcolor{teal}{400 billion per annum} as of 2021.

[\textbf{Evidence2}]: Diaspora remittances were robust in 2021, with total \textcolor{teal}{remittances in 2021 reaching a record USD $3,718$ million}. of daily users

 [\textbf{Verdict}]:\textcolor{teal}{\textsf{True}} \\

\textbf{LLM reasoning without \name{}} \\
\textbf{[Reasoning]:} The evidence \textcolor{red}{provides conflicting information} about the exact amount of \textcolor{red}{diaspora remittances} in 2021, and some sources mention figures that \textcolor{red}{are not directly comparable to KSh400 billion.}

\textbf{[Prediction]:} \textcolor{red}{Conflicting} \\

\textbf{LLM reasoning  with \name{}} 

 \textbf{[Reasoning]:}       The evidence supports the claim that diaspora remittances reached an all-time high of over KSh400 billion in 2021. However, \textcolor{teal}{there are minor variations in the wording and units used across different sources.}

 \paragraph{[\textbf{Prediction}]:}\textcolor{teal}{\textsf{True}}
\end{tcolorbox}

\vspace{-1em}
\captionof{figure}{Example numerical claim demonstrating \textit{\analysisissue} issue and mitigation offered by \name{}}\label{intro:example} 
\end{table}
\section{Introduction}
\label{chapter:introduction}

Automated verification of claims is critical to mitigate misinformation and disinformation. While many approaches focus on synthetically generated claims from Wikipedia \cite{hover,fever,feverous}, recent works have focused on fact-checking real-world claims \cite{multifc,averitec,claimdecomp,quantemp}. 

Real-world claims contain numerical (quantitative and temporal) elements \cite{quantemp,alam2025clef} and verifying such claims necessitates systematically comparing numerical aspects within the claim against corresponding evidence provided. Consider the example in Figure \ref{intro:example}, where verifying the claim about diaspora remittances requires ensuring that the claimed amount (KSh400 billion) and the specified year (2021) align correctly with the evidence, even when the evidence presents the amount in a different currency (USD). Unlike binary verification tasks, real-world claims often contain contradictory or partially correct components, making verification more complex.

Standard automated fact-checking approaches employ a multistage pipeline involving fine-tuned NLI models. Recently, Large Language Models (LLMs) have emerged as promising alternatives to NLI-based methods for claim verification \cite{self_checker}. However, these approaches have been evaluated mainly in synthetic claims, and recent studies highlight significant shortcomings when applied to complex claims in the real world that involve numerical data \cite{quantemp,llms_low_perf, averitec,augenstein2024factuality}. Although prior work has acknowledged these limitations, they have not thoroughly analyzed why LLMs fail in realistic scenarios. 

To address this gap, we analyze failure modes in both open-source and proprietary models of different sizes. Beyond noise in retrieval, we identify a critical reasoning challenge even in the presence of relevant evidence introduced specifically due to multi-aspect or inability to contextualize diverse information from evidence sources, termed \textbf{\analysisissue} due to issues in its ability for accurate analysis of claims and evidence. This differs fundamentally from the ``overthinking'' phenomenon observed in simpler Q\&A contexts, where models unnecessarily elaborate on obvious details \cite{sui2025stopoverthinkingsurveyefficient}. While it can impact simple claims, \analysisissue~is especially pronounced in complex cases with multiple aspects and diverse evidences, leading to unfocused reasoning that strays from the core verification task. Our analysis reveals that approximately \textbf{34\%} of claims in the QuanTemp dataset suffer from this reasoning problem.

In the context of fact-checking, \textit{\Analysisissue} stems from LLMs focusing on unimportant aspects of the claim or misinterpretation of information in diverse evidences as part of its reasoning process. For instance, in the example in Figure \ref{intro:example} the claim mentions an amount of \textit{400 billion Ksh} (kenyan shillings). While one of the evidences confirm this figure, the other evidence shown in the example quotes an amount of 3718 mil. USD. Though this is equivalent to 400 billion ksh, the LLM interprets this quantity as not being equivalent to the one mentioned in the claim and falls short by billions, predicting its verdict as ``Conflicting".  To our knowledge, our work is the first to study the problem \textit{\analysisissue} occuring due to \textbf{overanalysis} and explore test-time scaling as a solution to the same in the context of fact checking of real-world claims. 

More formally \analysisissue \ due to overanalysis can be defined as the phenomenon in which:

\emph{``The LLM backtracks from it’s current accurate reasoning trajectory due to misinterpretation of diverse information which may also stem from multi-aspect nature of claims and inability to contextualize information."}

We introduce and investigate test-time scaling (TTS)~\cite{zhang2025survey}, an approach that allocates additional computational resources at inference time, enabling LLMs to explore multiple reasoning pathways to mitigate \analysisissue. To our knowledge, this is the first study to examine \textbf{test-time scaling} specifically for verifying complex factual claims. To effectively navigate and select from diverse reasoning paths, we train a dedicated verifier model (\name{}). Our experiments demonstrate that employing this verifier with a \textbf{best-of-N (BoN)} strategy significantly outperforms standard \textit{majority-voting} \cite{self_consistency} methods. 

Since real-world claims are diverse, applying a fixed uniform computational effort across all claims is inefficient: simpler claims may consume unnecessary resources, and complex claims may not receive sufficient computational attention, resulting in suboptimal verification. To address this issue, we propose a novel \textit{adaptive TTS} strategy that \textit{dynamically} adjusts computational resources based on claim complexity, determined by analyzing \textit{layerwise latent representations} within the LLMs. This adaptive method boosts verification performance by \textbf{18.8\%} on complex numerical claims (QuanTemp) and \textbf{21.27\%} on compositional claims (ClaimDecomp) compared to single-shot techniques, while also delivering \textbf{1.8x} greater computational efficiency than the standard TTS approach.

In this paper, we answer the following research questions through empirical validation:

\textbf{RQ1}: To what extent LLMs suffer from \textit{\analysisissue} issue for complex numerical fact-checking tasks?

\textbf{RQ2}: How does test-time scaling impact the reasoning performance of LLMs for fact-checking complex numerical claims?

\textbf{RQ3}: How does test-time scaling compare to current strategies for  fact-checking complex claims?

\section{Related Work}
\subsection{Fact Checking Complex Real-World Claims}

Automated fact-checking is key to mitigating growing  misinformation and disinformation \cite{fever,feverous,guo2022survey}. Existing works on automated fact-checking primarily focus on synthetic claims collected from Wikipedia \cite{fever,hover,feverous,nanekhan2025flashcheck} that are not representative of real-world claims.  More efforts have been made to build systems for real-world claims in domains like politics \cite{claimdecomp,liar,political_claims_1,multi_hop_political,nakov2025overview}, science \cite{wadden-etal-2020-fact,vladika2023scientific,wright2022generating}, health \cite{kotonya2020explainable} and climate \cite{diggelmann2021climatefever}.

\subsection{Reasoning drift due to overthinking in LLMs}

LLMs with a large number of parameters are believed to exhibit emergent capabilities \cite{wei2022emergent,brown2020gpt3,openai2023gpt4}, including the ability to reason. Prior work shows that, when given few-shot examples, LLMs can produce step-by-step explanations (known as chains of thought or reasoning chains), resulting in improved performance on reasoning tasks \cite{wei2023chainofthought,huang-chang-2023-towards,yao2023react,huang-chang-2023-towards}. However, there are known blindspots when used in few-shot and one-shot settings especially when numerical, temporal data is involved~\cite{wallat2024study,wallat2024temporal}.

Moreover, recent studies show that large language models (LLMs) are prone to \emph{overthinking} when they perform reasoning, a phenomenon where models generate unnecessarily long or elaborate reasoning chains that ultimately lead to incorrect conclusions \cite{sui2025stopoverthinkingsurveyefficient,chen2025think23overthinkingo1like}. Overthinking has also been described as the tendency of LLMs to prioritize internal reasoning over external feedback \cite{cuadron2025dangeroverthinkingexaminingreasoningaction}. In knowledge-intensive tasks such as fact-checking, we observe that LLMs often backtrack or \textbf{overanalyze} their initial reasoning chain. This typically arises from the complexity of queries involving multiple aspects, or from misinterpreting conflicting information in the evidence. Crucially, this reasoning gap can occur regardless of the actual length of the reasoning chain. We refer to this problem as \textbf{\analysisissue}. While prior work has explored the application of LLMs to fact-checking \cite{self_checker}, none have explicitly addressed the \analysisissue\ phenomenon. Our findings suggest that \analysisissue\ presents a significant challenge for verifying complex numerical claims.

\subsection{Test Time Scaling}

Test-Time Scaling is a technique to improve the reasoning performance of LLMs by allocating additional inference-time compute~\cite{zhang2025survey,venktesh2025trust:verifiers:survey}. Early approaches primarily focused on generating multiple outputs per query in parallel and selecting an answer via majority voting \cite{self_consistency} or search-based methods \cite{tree_of_thought, alpha-zero-like-tree-search, self-evaluation-beam-search}. Recent developments can be broadly categorized into two lines of work: (i) modifying the proposal distribution of LLM by fine-tuning \cite{deepseek-r1, o1-paper, singh2024humandatascalingselftraining,
NEURIPS2023_91edff07, zelikman2022starbootstrappingreasoningreasoning, search-r1}, and (ii) training a reward / verifier model to select the preferred answer from multiple candidates \cite{snell_TTS, math_sheperd, nichols_verifier, cobbe_2021_ORM, uesato_PRM, ligthman_PRM}. Our methodology focuses on the latter, often referred to as \textit{search against a verifier.} It involves sampling multiple outputs and using a trained verifier model to select the best final output or the reasoning path that led to it. 

Verifiers used to select the best solution, are divided into Outcome reward models (ORM) \cite{cobbe_2021_ORM} or Process Reward Models (PRM) \cite{uesato_PRM,ligthman_PRM}. We adopt a PRM based verifier and while the majority of the approaches use rejection sampling training only on positive reasoning paths, we also include incorrect solutions to provide better discriminative capacity to the verifier. Existing works also apply TTS uniformly to all test sample, while some claims may not require TTS, wasting compute resources. We are the first to explore applying TTS adaptively based on complexity level of the claim, drawing inspiration from adaptive RAG \cite{adaptive_RAG}.

\section{Methodology}
\label{sec:methods}
\begin{figure*}[h]
    \centering
    \fbox{\includegraphics[width=0.87\linewidth]{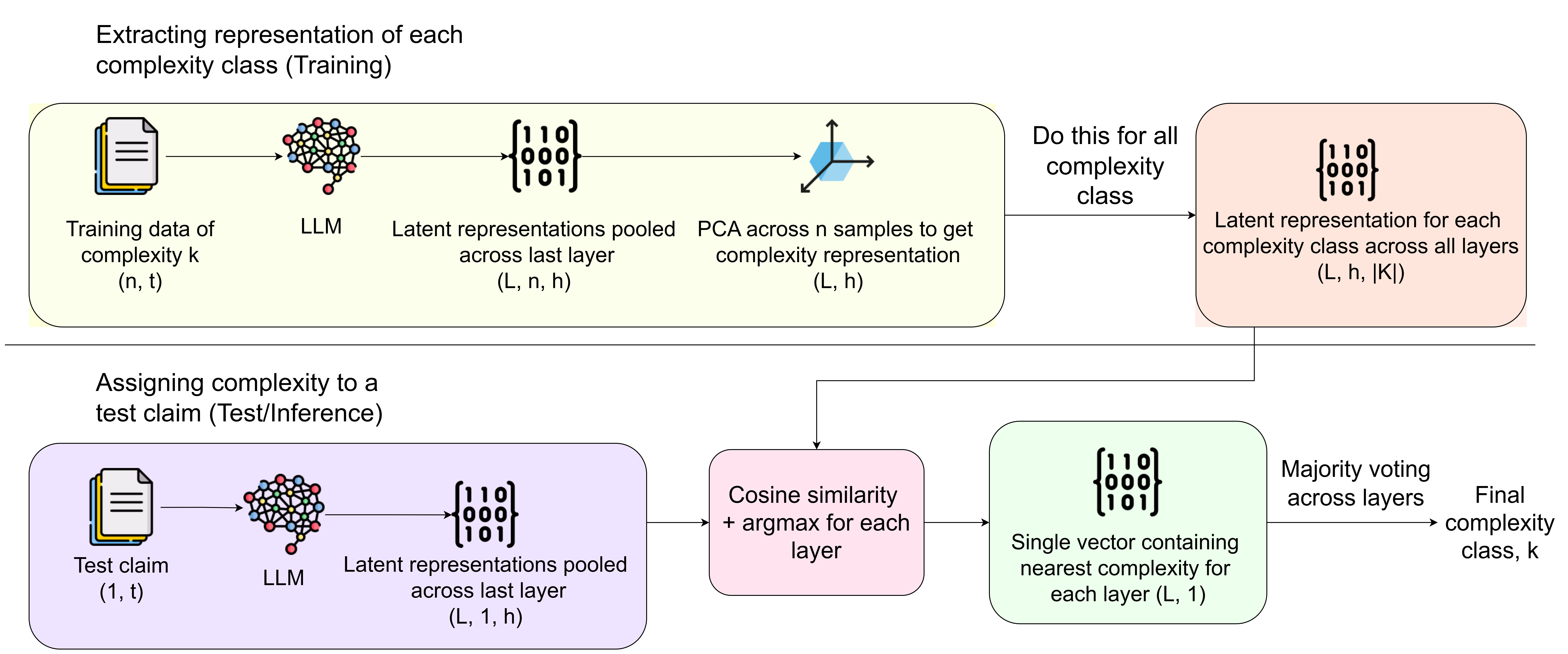}}
    \caption{Training and inference of complexity assignment pipeline using latent representation-based method for adaptive decomposition.}
    \label{fig:ctrla_pipeline}
\end{figure*}
In this section, we present a test-time scaling (TTS) strategy aimed at mitigating \analysisissue, a reasoning failure wherein the model fails to prioritize relevant information or misconstrues aspects of the claim during inference. Our approach involves training a process verifier (Section \ref{sec:verifier}) to select more accurate reasoning chains from multiple candidates. To further reduce unnecessary compute on straightforward claims, we explore an adaptive TTS mechanism that predicts claim complexity using layer-wise latent representations from the base LLM (Section \ref{sec:adaptive}). We adopt a standard fact-checking pipeline of BM25 retrieval + re-ranking to extract evidences for claim and employ LLMs for claim verification instead of standard NLI models.


\subsection{Verifier for best path selection in TTS}
\label{sec:verifier}

Test-time scaling using a best-of-$N$ strategy, in the context of fact-checking, entails generating multiple candidate reasoning paths for verifying a claim $c$, followed by selecting the best verdict using a verifier $\mathcal{V}$. We propose training a \textit{process-based verification model} that scores reasoning paths rather than just final verdicts for a given claim, enabling the selection of the most insightful path for verification.

More formally, given a training corpus $D$ of claims $(c_i)$ and corresponding expert fact-checker verdicts $(y_i)$, where $D = \{(c_1, y_1), \dots, (c_n, y_n)\}$, we first sample $m$ different reasoning paths $\{(e_{11}, \dots, e_{1m}), \dots, (e_{n1}, \dots, e_{nm})\}$ from a pre-trained LLM $\mathcal{F}_{\text{LLM}}$ for each claim during the claim verification phase, along with their corresponding predicted verdicts $\{(\hat{y}_{11}, \dots, \hat{y}_{1m}), \dots, (\hat{y}_{n1}, \dots, \hat{y}_{nm})\}$. We then generate training data for the verifier by labelling each $(c_i, e_{ij}, \hat{y}_{ij})$ pair as 1 (positive sample) if the predicted verdict matches the ground-truth ($\hat{y}_{ij} == y_i$), and 0 otherwise (negative sample).

We use this constructed dataset of positive and negative reasoning paths to train the verifier model $\mathcal{V}(sc_i \mid (c_i, e_{ij}, \hat{y}_{ij}))$, which takes as input the original claim $c_i$, a reasoning path $e_{ij}$, and its corresponding predicted verdict $\hat{y}_{ij}$, and outputs a score based on the plausibility of the path leading to the correct verdict. The model is fine-tuned to assign higher scores to more appropriate reasoning paths by minimising a binary cross-entropy loss:
\begin{align*}
\mathcal{L}_{\mathcal{V}} = & -[ y_i \cdot \log \mathcal{V}(sc_i \mid (c_i, e_{ij}, \hat{y}_{ij})) \\
& + (1 - y_i) \cdot \log (1 - \mathcal{V}(sc_i \mid (c_i, e_{ij}, \hat{y}_{ij}))) ]
\end{align*}
The existing process verification approaches for training verifier primarily focus on rejection sampling when generating training data where only positive samples are considered. However, the supervision from training data generated above, helps the verifier account for nuances in the claim verification process by learning the ability to distinguish better between plausible and incorrect paths.

During inference for a test claim $c_i$ we generate $m$ such reasoning paths $\{(e_{11}, \hat{y}_{11})...(e_{1m}, \hat{y}_{1m}))\}$  and employ the verifier to score these reasoning paths by extracting logits score from output of the verifier  $\cV(sc_i | (c_i,e_{ij},\hat{y_{ij})} .. )$. The path with highest score is given as outcome of claim verification process. However, this best of N (BoN) strategy applies TTS for all claims uniformly not accounting for a scenario where TTS might not be useful. This is not compute-optimal and we propose an adaptive BoN strategy to circumvent this issue.

\subsection{Adaptive Test-Time Scaling (TTS)}
\label{sec:adaptive}

 We adopt a semi-supervised approach to estimate the \textit{complexity} of a claim by leveraging sub-question decomposition and clustering in the latent representation space of a language model. Our method is inspired by recent work on Adaptive RAG \cite{adaptive_RAG,mallen_adaptive_RAG}, which uses query complexity annotations to guide inference-time control in retrieval-augmented generation. In our case, we define complexity whether the claim requires decomposition to correctly verify its veracity. This decomposition signal serves as a proxy for reasoning difficulty, which we use to determine whether a claim should \textit{trigger TTS}.

We construct the training data for adaptive TTS by annotating claims with sub-questions and track the number required to resolve the claim’s veracity correctly. Claims are then categorized into two levels of complexity: level 0 includes claims that can be correctly verified without any decomposition; level 1 includes claims that require one or more sub-questions and are correctly verified after decomposition; and claims that remain misclassified even after decomposition are also assigned to level 1, reflecting unresolved ambiguity or model uncertainty.





Our adaptive decomposition decision algorithm draws inspiration from CTRLA \cite{ctrlA}, which posits that LLM latent representations encode semantically meaningful attributes. As shown in Figure~\ref{fig:ctrla_pipeline}, we extract latent representations and assign complexity labels accordingly. 

We annotate a training set of claims \( D_k = \{c_1, c_2, \dots, c_n\} \) with complexity level \( k \). Each claim is passed through a pretrained LLM \( \mathcal{F}_{\text{LLM}} \), with \( L \) layers and latent size \( h \), to extract token-level latent representations, following prior work on semantic properties of latent representations \cite{latentRep_sun2025, latentRep_chuang2024}. For an input claim \( c_i \) with \( t \) tokens, its latent representation $\mathbf{r^l}$ for layer $l$ is defined as:
\begin{align*}
\mathbf{r}^l &= \mathcal{F}^{l}_{\text{LLM}}(c_i) \\
              &= [\mathbf{r}^{(l)}_{i,1}, \mathbf{r}^{(l)}_{i,2}, \dots, \mathbf{r}^{(l)}_{i,t}] \in \mathbb{R}^{t \times h}
\end{align*}
More generally, the set of token representations across all layers can be expressed as:
\begin{align*}
\mathcal{F}_{\text{LLM}}(c_i) = \left\{ \mathbf{r}_{i,j}^l \in \mathbb{R}^h \,\middle|\, 1 \leq j \leq t,\ 1 \leq l \leq L \right\},
\end{align*}
where \( \mathbf{r}_{i,j}^l \) is the hidden representation of the \( j^\text{th} \) token at layer \( l \). Then, we summarize this representation using the last token as a proxy for the entire claim (following \cite{attention_all_you_need}). Formally, the pooled representation at layer $l$ is defined as: 
\begin{align*}
\mathbf{z}_i^l = \mathcal{P}(\{\mathbf{r}_{i,j}^l\}_{j=1}^t) = \mathbf{r}_{i,t}^l \quad \text{for } 1 \leq l \leq L,
\end{align*}
where $\mathcal{P}$ denotes the pooling function over the token dimension. Generally, this can be represented as:
\begin{align*}
\mathcal{Z}_i = \left\{ \mathbf{z}_i^l \in \mathbb{R}^h \,\middle|\, 1 \leq l \leq L \right\}.
\end{align*}

We then apply Principal Component Analysis (PCA) to \( \left\{ \mathbf{z}_i^l \right\}_{i=1}^n \) for each layer \( l \), and extract the first principal component:
\begin{align*}
\mathbf{u}_k^l = \text{PCA}_1\left( \left\{ \mathbf{z}_i^l \right\}_{i=1}^n \right),
\end{align*}
yielding complexity level representations: \begin{align*}
\mathcal{U}_k = \left\{ \mathbf{u}_k^l \right\}_{l=1}^L,\quad \\
\end{align*}

For a test claim \( c \), we extract its layerwise representations:
\begin{align*}
\mathcal{C} = \left\{ \mathbf{c}^l \in \mathbb{R}^h \,\middle|\, 1 \leq l \leq L \right\}.
\end{align*}
Cosine similarity is computed with each class prototype at each layer:
\begin{align*}
\mathcal{S} = \left\{ \mathbf{s}^l \in \mathbb{R}^{|K|} \,\middle|\, 
\mathbf{s}_k^l = \cos\left( \mathbf{c}^l, \mathbf{u}_k^l \right), \right. \\
\left. 1 \leq k \leq |K|,\ 1 \leq l \leq L \right\}
\end{align*}
where \( \mathbf{s}_k^l \) is the similarity with class \( k \) at layer \( l \). We assign:\[
\hat{k}^l = \arg\max_k\ \mathbf{s}_k^l
\]and the final predicted complexity class is determined via majority voting across \( \{\hat{k}^l\}_{l=1}^L \).
This approach is similar to prototypical classification in ML literature \cite{prototypical_networks,prototypical_networks_2} or nearest centroid / rocchio classifier \cite{rocchio71relevance}. The approach primarily pre-computes and indexes latent representations for complexity classes. Then at inference time we extract latent representations for the test claim and assign the complexity class based on the representation that is closest as measured by cosine similarity, similar to Rocchio classifier in IR literature.
Based on the inferred complexity, we apply adaptive test-time scaling: one-shot inference for level-0 claims, and expanded reasoning for level-1 (moderate or high complexity) claims. This yields a compute-efficient TTS strategy for complex fact-checking, as shown in Section~\ref{sec:results}.

\section{Experimental Setup}
 \label{sec:experiments}

 \textbf{Datasets.}
We evaluate our methods on  \textbf{QuanTemp} \cite{quantemp}, which is the first benchmark for complex numerical claims. It comprises 9935 training, 3084 validation and 2495 test claims. We employ their training set to train the verifier model and create representations for the complexity classes, which are used for adaptive TTS. 

\textbf{ClaimDecomp}~\cite{claimdecomp} is another benchmark dataset that focuses on compositional real-world claims. We use its evaluation set (200 claims) to test the generalization capabilities of our verifier (trained on QuanTemp) to other types of complex claims.

\label{sec:results}
\begin{table*}[htb]
\centering
\resizebox{\textwidth}{!}{
\begin{tabular}{lcccccccccc}
\toprule
\multirow{2}{*}{\textbf{Method}}  & \multicolumn{5}{c}{\textbf{QuanTemp}} & \multicolumn{5}{c}{\textbf{ClaimDecomp}}\\
\cmidrule(lr){2-6} \cmidrule(lr){7-11}
  & T-F1 & F-F1 & C-F1 & \textbf{M-F1} & \textbf{W-F1}  & T-F1 & F-F1 & C-F1 & \textbf{M-F1} & \textbf{W-F1} \\
\midrule

\textbf{Other Baselines} \\
DeepSeek-R1 (7B) \cite{deepseek-r1}  & 39.40 & 58.80 & 28.80  & 42.3 & 47.9  & 36.89 & 52.56 & 25.29 & 38.25 & 41.21\\
\textbf{Llama-3.1 8B} \\

Top 1 Decoding \cite{wei2023chainofthought} & 33.5 & 66.4 & 34.7 & 44.8  & 52.5 & 19.75 & 44.03 & 44.44 & 36.07 & 37.15  \\

Self-Consistency \cite{self_consistency} & 
34.9 & 68.6 & 35.1 & 
46.2 & 
54.2  & 16.22 & 44.00 & 44.30 & 34.84 & 36.39\\

Best of N  & 
 44.20 & 75.30 & 40.30 &
53.20  & 
\textbf{61.0} & 22.22 & 48.00 & 46.91 & 39.05 & 40.22  \\
Adaptive BoN (\name{}) & 44.27 & 73.52 & \textbf{43.93} & \textbf{53.91} &60.84 & 21.43 & 56.21 & \textbf{49.70} & \textbf{42.44} & \textbf{44.13} \\ 
\hline
 Upper bound & 63.9 &  85.1 & 69.60 & 72.9  & 77.4 & 63.92 & 82.21 & 83.10 & 76.41 & 77.21 \\
\midrule
\textbf{Gpt-4o-mini} \\
Top 1 decoding \cite{wei2023chainofthought} & 40.68 &63.14 & 37.99 &47.27 & 52.84 & 42.11 & 48.92 & 39.06 & 43.36 & 44.46 \\
Self-Consistency \cite{self_consistency} & 29.70 & 63.20& 42.33 &  45.07 & 51.83  & 47.41 & 50.00 & 34.19 & 43.87 & 45.35 \\
Best of N  & 26.30 & 74.82  & 46.51 & 49.21& 58.81 & 45.45  &  59.60 & 39.32 & 48.12 & 50.39  \\
Adaptive BoN (\name{})& \textbf{36.16}& 72.61  & 44.22 & \textbf{51.00} & \textbf{58.88} & 49.18 & 57.93 & \textbf{46.62} & \textbf{51.24} & \textbf{52.53} \\ 
\hline
 Upper bound & 45.34 & 85.37 & 67.19 &  65.97 & 73.41 & 58.21 & 66.67& 61.68 & 62.19 & 62.91 \\ \hline
\textbf{Oracle (fine-tuned)} \cite{quantemp} & 56.86 & 82.92 & 48.79 & 62.85 & 69.79 & 66.67 & 66.67 & 45.76 & 59.69 & 60.50 \\

\bottomrule
\end{tabular}}

\captionsetup{font=small}
\caption{Claim verification performance of the models. The reward model is fine-tuned llama-3.2-3b and number of reasoning paths considered, m=10. The gains method is computed w.r.t the \textit{top 1 Decoding} baseline.}
\label{tab:rm_performance_table}
\end{table*}

\begin{table}[ht]
\small
\begin{tabular}{lcccccc}
\toprule
  \textbf{Decomposition} & \textbf{T-F1} & \textbf{F-F1} & \textbf{C-F1} & \textbf{M-F1} & \textbf{W-F1} \\
\midrule

\multicolumn{6}{l}{\textbf{Majority Voting}} \\

\cmidrule(lr){1-6}
 w/o decomp      & 34.89 & 68.63 & 35.09 & 46.20 & 54.18 \\
 with decomp  & 37.48 & 69.33 & 32.27 & 46.36 & 53.51 \\
\midrule

\multicolumn{6}{l}{\textbf{BoN }} \\
\cmidrule(lr){1-6}
 w/o decomp     & \textbf{44.16} & 75.27 & \textbf{40.27} & \textbf{53.23} & \textbf{60.97} \\
 with decomp   & 43.77 & \textbf{76.32} & 35.21 & 51.77 & 60.29 \\
\midrule



\end{tabular}
\caption{Analysis on impact of claim decomposition in test-time scaling setup on complex claim verification (using Llama-3.1 8B as base model).}
\label{tab:decomp_analysis}
\end{table}
\noindent \textbf{Fact-checking pipeline}: We use a standard fact-checking pipeline as adopted in \cite{quantemp} which comprises an initial retrieval of top-100 results for each claim using BM25 (Elasticsearch). We then re-rank using paraphrase-MiniLM-L6-v2 \cite{sentence-transformers} to select top-3 evidence snippets for each claim. \\
\noindent \textbf{Metrics}: We use per-class F1 scores, macro and weighted F1 score as evaluation metrics.\\
 \noindent \textbf{LLMs used for claim verification}: 
We use Llama 3.1 (8B) and gpt-4o-mini as LLMs for claim verification stage and Deepseek-r1 which internally does TTS as baseline. We employ a temperature of 0.45 for generating diverse reasoning paths in TTS setting. We use Few-shot Chain of Thought (COT) prompting (in \textbf{Figure \ref{fig:fact_checking_prompt}}) for eliciting reasoning for all baselines and our approach. We experiment with different number of reasoning paths (m=1,3,5,10) as elaborated in Section \ref{sec:results}.
 
 \noindent \textbf{Verifier model for TTS}: The Verifier model used for Best-of-N in TTS setting is based on the \texttt{Llama-3.2-3B} model. It is fine-tuned for classification using LoRA adapters~\cite{lora-peft} with a rank of 8 and a scaling factor (\(\alpha\)) of 16. The model is fine-tuned for 3 epochs with a batch size of 32, using the AdamW optimizer with \(\epsilon = 1 \times 10^{-8}\) and an initial learning rate of \(1 \times 10^{-3}\).

 \textbf{Prompts}: All prompts for claim verification, decomposition and LLM based annotations can be found in Appendix \ref{appendix:prompts}.

\section{Experimental Results}
\label{sec:results}

\begin{table*}[hbt!]
\small
    \setlength{\tabcolsep}{1.5pt}
    \begin{tabular}{p{2.5cm}p{13cm}}
\toprule
    \textbf{Method}     & \textbf{Evidences} \\
\midrule

\midrule

\colorg \textbf{Claim}  & \colorg \textbf{       Diaspora remittances “now stand at an all-time high of over KSh400 billion in 2021”.}  \\
  & [Evidence 1]:  diaspora remittances have increased tenfold over the last decade; and they now stand at an all-time-high at over ksh. \textcolor{teal}{400 billion per annum} as of 2021.  \\
& [Evidence 2]:  diaspora remittances were robust in 2021, with total \textcolor{teal}{remittances in 2021 reaching a record usd3,718 million}. of daily users. \\
Top-1 decoding & 
[Reasoning]:  The evidence \textcolor{red}{provides conflicting information about the exact amount of diaspora remittances in 2021}, and \textcolor{red}{some sources mention figures that are not directly comparable to KSh400 billion.}
          \\
& [Verdict]: \textcolor{red}{Conflicting} \\
Self Consistency & [Final Verdict]: \textcolor{red}{Conflicting} \\
\\
 \name{} (ours) 
& [Reasoning]:          The evidence supports the claim that diaspora remittances reached an all-time high of over KSh400 billion in 2021. However, \textcolor{teal}{there are minor variations in the wording and units used across different sources.} \\
& [Final Verdict]: \textcolor{teal}{True}
 \\

\midrule

\colorg \textbf{Claim}  & \colorg \textbf{The 1956 Republican Party platform supported equal pay, the minimum wage, asylum for refugees, protections for unions and more.}  \\
 & [Evidence 1]:  1928 the gop platform read: the party favors freedom in wage we support an equitable minimum wage for american workers one providing fair wages  \\
& [Evidence 2]:  the 1956 republican platform pledged to provide assistance; protect social security; provide  \textcolor{teal}{asylum for refugees; extend the minimum wage;}  \textcolor{teal}{strengthen labor laws to support union; assure equal pay...} \\
Top-1 decoding & [Reasoning]:        The evidence provided supports many of the claims made in the original statement, \textcolor{red}{but some quotes are from earlier years or mention specific policies}, and not all points are explicitly stated in the 1956 platform. \\
& [Final Verdict]: \textcolor{red}{Conflicting} \\

Self Consistency & [Final Verdict]: \textcolor{red}{Conflicting} \\
\\
  & [Reasoning]: The claim is supported by quotes from the 1956 Republican platform, which mention \textcolor{teal}{equal pay, minimum wage, asylum for refugees, protections for unions,}. While there are \textcolor{teal}{minor nuances} it is supported [\dots] \\
 \name{} (ours)  & [Final Verdict]: \textcolor{teal}{True}\\
\midrule

\end{tabular}
\caption{Qualitative analysis of reasoning process by Self consistency and \name{}}
\label{tab:qualitative}
\end{table*}

\subsection{LLM Reasoning Drift - Manual Analysis}
To answer \textbf{RQ1}, we conduct a systematic study of \textit{\analysisissue} in numerical claim verification using QuanTemp. From a stratified sample of 100 claims, we use a system prompt (Figure~\ref{tab:system_prompt_llm_judge}) and a few-shot prompt with annotated examples (Figure~\ref{tab:overanalysis_1}, Appendix~\ref{appendix:prompts}) to instruct an LLM judge (gpt-4o-mini) to label each reasoning path as exhibiting \textit{\analysisissue} (1) or not (0).

Two human annotators (graduate volunteers) then independently verify whether the LLM's label is correct, following the guidelines in Appendix~\ref{appendix:annotations}. We report a Cohen’s Kappa of \textbf{74.13} indicating substantial agreement and LLM judge is \textbf{69\%} accurate. Based on this procedure, after correcting the LLM judge outcomes, we find that approximately \textbf{34\%} of sampled claims exhibit \textit{\analysisissue}, suggesting it is a common and non-trivial failure mode. Examples are shown in Figure~\ref{intro:example}, Table~\ref{tab:qualitative}, and Figures~\ref{tab:overanalysis_example1},~\ref{tab:overanalysis_example2} (Appendix~\ref{sec:overanalysis_example}).

\subsection{Effectiveness of test-time scaling for Claim Verification}

To address \textbf{RQ2}, we evaluate TTS as a mitigation strategy (Section~\ref{sec:methods}) with results shown in Table~\ref{tab:rm_performance_table}. We first establish a theoretical upper bound, where a perfect verifier selects the correct reasoning path and verdict whenever present.

As shown in Table~\ref{tab:rm_performance_table}, the upper bound achievable via TTS—across different LLMs—surpasses the oracle performance of fine-tuned NLI models reported in \citet{quantemp}. The \textit{Oracle fine-tuned} row reflects state-of-the-art NLI models (e.g., RoBERTa-large) trained on human-written justifications and provided those at inference. In contrast, the TTS upper bound relies entirely on LLM-generated reasoning, without any human-authored justifications. This highlights LLMs' ability to elicit appropriate reasoning processes for complex fact-checking when guided by TTS. 

Self-consistency \cite{self_consistency}, which aggregates multiple reasoning paths via majority voting, performs only marginally better than—or sometimes worse than—top-1 decoding. This is because it assumes the most frequent verdict is correct, overlooking the nuanced reasoning often required in fact-checking, especially for conflicting claims with diverse evidence.

As shown in Table~\ref{tab:rm_performance_table}, our trained verifier in both BoN and Adaptive BoN settings effectively selects the best reasoning path by leveraging task-specific signals. Adaptive BoN, which invokes TTS based on the predicted claim complexity, improves over top-1 decoding by \textbf{18.8\%} and self-consistency by \textbf{16.69\%} (Macro-F1 on QuanTemp). Adaptive BoN also outperforms \textbf{DeepSeek-R1} by \textbf{27.44\%}, highlighting the importance of a trained verifier for TTS in complex fact-checking.

\begin{figure}[t!!!]
    \centering
    \begin{subfigure}{.45\linewidth}
\begin{tikzpicture}

    \begin{axis}[
        ybar=0.2pt,
            width=4cm,
            height=4.2cm,
            bar width=0.25,
            bar shift=0pt,
            every axis plot/.append style={fill},
            grid=major,
            xticklabel style={rotate=10},
            xtick={1, 2},
            xticklabels={\name{}, BoN},
            xlabel={},
            ylabel style = {font= \tiny},
        yticklabel style = {font=\boldmath \tiny, xshift=0.2ex},
        xticklabel style ={font=\small , yshift=0.2ex},
            ylabel={Avg runtime per query },
            enlarge x limits=0.25,
            ymin=0,
            ymax=9200,
            legend style ={font=\small,yshift=0.5ex},
            area legend,
            nodes near coords style={font=\tiny,align=center,text width=2em},
            legend pos=north west,
            legend columns=-1,
            legend style={/tikz/every even column/.append style={column sep=0.2cm}},
        ]
        \addplot+[
            ybar=0.2pt,
            plotColor1*,
            draw=black,
            error bars/.cd,
                y dir=both,
                y explicit,
        ] plot coordinates {
                (1,4970) +- (0,0.38)
            };
            \addplot+[           ybar,
            plotColor2*,
            draw=black,
            postaction={
                    pattern=north east lines
                },
            error bars/.cd,
                y dir=both,
                y explicit,] plot coordinates {           (2,8500) +- (0,0.16)};

    \end{axis}
\end{tikzpicture}
\end{subfigure}
\hspace{0.3em}
    \begin{subfigure}{.45\linewidth}
\begin{tikzpicture}

    \begin{axis}[
        ybar=0.2pt,
            width=4cm,
            height=4.2cm,
            bar width=0.25,
            bar shift=0pt,
            every axis plot/.append style={fill},
            grid=major,
            xticklabel style={rotate=10},
            xtick={1, 2},
            xticklabels={\name{}, BoN},
            xlabel={},
            ylabel style = {font= \tiny},
        yticklabel style = {font=\boldmath \tiny, xshift=0.2ex},
        xticklabel style ={font=\small , yshift=0.2ex},
            ylabel={Total runtime per query (in minutes) },
            enlarge x limits=0.25,
            ymin=0,
            ymax=450,
            legend style ={font=\small,yshift=0.5ex},
            area legend,
            nodes near coords style={font=\tiny,align=center,text width=2em},
            legend pos=north west,
            legend columns=-1,
            legend style={/tikz/every even column/.append style={column sep=0.2cm}},
        ]
        \addplot+[
            ybar=0.2pt,
            plotColor1*,
            draw=black,
            error bars/.cd,
                y dir=both,
                y explicit,
        ] plot coordinates {
                (1,206.53)
            };
            \addplot+[           ybar,
            plotColor2*,
            draw=black,
            postaction={
                    pattern=north east lines
                },
            error bars/.cd,
                y dir=both,
                y explicit,] plot coordinates {           (2,349.59) };

    \end{axis}
\end{tikzpicture}
\end{subfigure}
\vspace{-2em}
\caption{(Left) Avg. runtime per query (in milliseconds) comparison and  (Right) Total \textbf{Runtime} comparison: Adaptive BoN (\name{}) vs BoN (Best of N).}
\label{fig:llm_calls}
\end{figure}
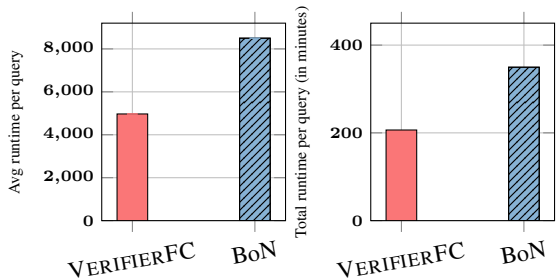
We observe similar improvements on Claimdecomp eval set where Adaptive BoN (\name{}) outperforms Top-1 decoding significantly, demonstrating that our verifier can also generalize to other types of complex claims apart from numerical claims. We observe that this is primarily because the verifier used in BoN and adaptive versions is trained on data that considers positive and negative reasoning paths / process paths from the LLM for claim verification. This is unlike conventional Process or Outcome Reward Models which employ rejection sampling and only consider positive samples. This offers more discriminative capacity to the verifier that is able to distinguish the correct reasoning path among diverse paths even when their frequency is low. For instance, on second example in Table \ref{tab:qualitative}, the LLM is unable to reason that similar party manifesto was also adopted earlier in 1928 and was further expanded in 1956 and hence the claim is accurate. Instead, the model overanalyzes, backtracks from correct reasoning and passes the verdict as ``Conflicting" due to mention of 1928 in one of the evidence not being in line with 1956 mentioned in claim. We observe that Self consistency also select ``Conflicting" as verdict as 9/10 reasoning paths lead to this verdict. Rather BoN and adaptive BoN was able to rank the one correct reasoning path as the best solution leading to correct verdict. 

We also observe that BoN and adaptive BoN with our trained process verifier helps mitigate the \analysisissue\ issue as evident from examples in Table \ref{tab:qualitative}. For instance, in the first example, one of the evidences exactly support the remittance value of 400 KSh and the second evidence quotes the equivalent value in US dollars. Under top-1-decoding, \analysisissue\ causes the LLM to classify the claim as "Conflicting" owing to the differing figures (ignoring the currency units) which is resolved when inferenced using BoN strategy (\name{}).


\subsection{Efficiency of Adaptive Test-Time Scaling}

We evaluate the efficiency of Adaptive BoN based TTS strategy by measuring total runtime on QuanTemp evaluation set and also record the average per claim processing time. The efficiency results are shown in plots in Figure \ref{fig:llm_calls} and claim verification performance in Table \ref{tab:rm_performance_table}. We observe that Adaptive Bon is less time intensive (\textbf{1.8$\times$}) as it adaptively scales inference computation based on the complexity of the claims. We also observe from the plots that Adaptive scaling of test-time compute also reduces overall running time significantly, while achieving gains over plain TTS based BoN strategy, leading to judicious use of computational resources without sacrificing performance. We observe that on QuanTemp most of the claims assigned complexity level 0 belong to \textbf{statistical} category which are easy to verify as noted by authors in \cite{quantemp}. We also independently evaluate the complexity consistency on its agreement with human notion of complexity by asking two annotators to annotate complexity levels for 100 claims sampled in a stratified manner. We observe a Cohen's kappa of \textbf{72.89} between annotators (Guidelines in Appendix \ref{appendix:guidelines}). Evaluating the classifier with annotations as ground truth, yields weighted precision of \textbf{71\%}.

\begin{table}[htb]
\centering
\small
\begin{tabular}{lcccc}
\toprule
\multirow{2}{*}{\textbf{Method}}  & \multicolumn{2}{c}{\textbf{QuanTemp}} & \multicolumn{2}{c}{\textbf{ClaimDecomp}}\\
\cmidrule(lr){2-3} \cmidrule(lr){3-5}
  &  \textbf{M-F1} & \textbf{W-F1}   & \textbf{M-F1} & \textbf{W-F1} \\
\midrule


\textbf{Llama-3.1 8B} \\

m=3  &50.60 & 57.47 & 37.96 & 39.71  \\
m=5 & 51.49 & 58.39 &39.64 & 41.25 \\
m=10 &  \bf 53.91 & \bf 60.84 &  \bf 42.44 &  \bf 44.13 \\
\midrule
\textbf{Gpt-4o-mini} \\
m=3 & 47.98 & 54.82 & 44.15 & 45.28 \\
m=5 & 49.26 & 56.82 & 48.24 & 49.24  \\
m=10 & \bf 51.00 & \bf 58.88 &  \bf 51.24 & \bf  52.53 \\

\bottomrule
\end{tabular}

\caption{Performance with different number of reasoning paths m=3,5,10.}
\label{tab:ablations}
\end{table}

\subsection{Impact of Decomposition on TTS}

Claim decomposition is a popular approach for addressing complex claims with multiple aspects \cite{quantemp,claimdecomp}. Hence, to answer \textbf{RQ3} we analyze the impact of decomposing claim into sub-questions in combination with test-time scaling to understand the impact on final claim verification. We use ClaimDecomp \cite{claimdecomp} style yes/no sub-questions as they are found to have superior performance and easy to comprehend \cite{quantemp,claimdecomp}. The results  are as shown in Table \ref{tab:decomp_analysis}. We observe that claim decomposition surprisingly leads to negative effects on claim verification performance. On analysis, we observe this is primarily due to \analysisissue\ issue from LLM being elicited due to certain sub-questions focusing on aspects that are not central to facts being checked, causing drift. 

Additionally, we observe that Adaptive BoN and BoN leads to diverse aspects of the claim being considered. Since this is the core objective of decomposition, TTS renders claim decomposition redundant, leading to diminishing returns (Tables \ref{tab:rm_performance_table}, \ref{tab:decomp_analysis} and \ref{tab:qualitative}). 



\subsection{Ablations}
We perform ablations by varying the scaling dimension of number of reasoning paths (m=3,5,10) using LLM. The results are as shown in Table \ref{tab:ablations}. We observe that the performance increases with increase in number of reasoning paths, reaching the best performance at m=10. We do not scale beyond 10 due to computational and cost constraints.

\section{Conclusion}


We identify and study the \textit{\analysisissue} issue in LLM reasoning for complex and numerical claims. To address this, we propose a Best-of-N (BoN) test-time scaling strategy, which outperforms Self-Consistency and Top-1 decoding. For improved compute-efficiency, we introduce Adaptive BoN (\name{}). \name{} achieves better performance than BoN while reducing computational cost. In future work, we aim to incorporate stronger verifier models and further optimize \name{} for performance and efficiency.

\section{Limitations}


While \name{} shows strong improvements over baselines, a notable gap remains compared to the upper bound. This may be addressed by integrating stronger verifier models with deeper linguistic and numerical understanding. Another promising direction is training verifiers on higher-quality, human-annotated data, which has been shown to improve reasoning performance. Additionally, our current approach leverages existing state-of-the-art retrieval and ranking approaches. However, noisy evidence can introduce bias that may affect both the performance and the validity of our analysis. Future work could incorporate more robust retrieval and re-ranking pipelines~\cite{rathee2024quam} to reduce this noise, enabling a fairer and more accurate evaluation of veracity prediction methods.

\section{Ethical Considerations and Risks}

This work focuses on improving factual consistency in language model outputs through test-time scaling and verifier-guided reasoning. While our methods aim to avoid \analysisissue, they are subject to several ethical considerations.

First, our reliance on LLM-generated reasoning paths introduces potential risks of reinforcing model biases or factual errors during verifier training. Although the verifier is trained to select plausible reasoning, it may inherit systematic flaws from the base LLM. Second, our approach does not currently include fairness or bias mitigation across demographic attributes, which may be relevant for politically or socially sensitive claims.

Our datasets (QuanTemp, ClaimDecomp) are drawn from public sources (with create commons license) and include no personally identifiable or sensitive information. The claims focus on public data and domain-specific knowledge rather than private individuals. Human annotators involved in evaluation were presented with reasoning paths and asked to assess model performance under clear, task-specific instructions (Appendix~\ref{appendix:annotations}). No potentially harmful or offensive content was encountered during the annotation process.

\section{Acknowledgements}

We would like to thank Sowmya AS for her help in evaluating quality of claim difficulty estimator. We would also like to acknowledge Factiverse AI for contributing hardware for the experiments reported in this work.



\bibliography{references}
\bibliographystyle{acl_natbib}

\clearpage
\appendix

\section{Annotation Guidelines}
\label{appendix:annotations}
\subsection{Guidelines for Manual analysis of \analysisissue\ issue}
Figure \ref{tab:overanalysis_annotator_guideline} presents the guidelines to the annotator qualitatively evaluating the samples identified by judge LLMs for \textit{\analysisissue}. Depending on the human judgements, they are instructed to assign a category \texttt{1} if they agree with the LLM's judgement and \texttt{0} otherwise. This assignment is used to compute Cohen's Kappa to evaluate inter-annotator agreements. 

\subsection{Annotation Guidelines for Claim Complexity}
\label{appendix:guidelines}
Figure \ref{fig:complexity} shows the guidelines for annotating complexity level for the claims.
\begin{table}
\begin{tcolorbox}[title=Annotator Guidelines for Claim Complexity Annotation]

\textbf{Annotation Instructions:}  

Please review the given claim in context of provided expert fact-checker justification and verdict. Trace the process of fact-checking by yourself and analyze how many aspects need to be verified in the claim. if the claim has more than one aspect that has to be verifier using multiple sources annotate it as 1 else annotate it as 0.
\end{tcolorbox}

\captionof{figure}{Annotator Guidelines for Claim Complexity Annotation.}
\label{fig:complexity}

\end{table}

\begin{table}[htbp]
\centering
\small
\begin{tcolorbox}[title=Annotator Guidelines for Evaluating Reasoning Drift]

Please review the LLM's judgment regarding the presence or absence of \analysisissue, along with its justification. Then, annotate whether you believe the label is correct and whether it aligns with our definition of \analysisissue.

\textbf{Definition of Reasoning Drift:}  
Reasoning Drift is defined as a behavior inherent to LLMs, where—particularly in the context of fact-checking—the model overfits to an aspect of the claim that is not central to verifying its truth. This can be triggered by irrelevant details or by diverse contexts.

\textbf{Examples:}
\begin{itemize}
    \item An LLM might fail to recognize that two quantities expressed in different units (e.g., KSh and dollars) represent the same value.
    \item It might not realize that the same political manifesto could have been adopted by a party in both 1928 and a later year, leading to unnecessary doubt.
\end{itemize}

\textbf{Annotation Instructions:}  
If the LLM judge has determined that \analysisissue\ is present (or absent), and you agree that this decision aligns with the definition above, mark it as \textbf{1} (i.e., correct). Otherwise, mark it as \textbf{0} (i.e., incorrect). We will use this to compute inter-annotator agreement.

\end{tcolorbox}

\captionof{figure}{Guidelines for human annotators evaluating the \analysisissue\ prediction made by the judge LLM.}
\label{tab:overanalysis_annotator_guideline}

\end{table}

\section{More examples exhibiting \analysisissue}
\label{sec:overanalysis_example}

Figure \ref{tab:overanalysis_example1} and Figure \ref{tab:overanalysis_example2} provides more qualitative examples where LLM exhibits \textit{\analysisissue} during reasoning. This is mitigated by selecting the \textit{best} reasoning sequence using \name{}.

\begin{table}[htb] 
\centering
\small
\begin{tcolorbox}[title=Mitigating \textit{\analysisissue} for Numerical FC]
\small
\medskip\noindent \paragraph{\textbf{Claim}:}\texttt{There's 100 million people that have preexisting conditions.}

 \paragraph{[\textbf{Evidence}]:}  a 2018 analysis by avalere, a health care consulting firm, \textcolor{red}{estimated that 102 million americans have preexisting conditions}. a 2019 analysis by the left-leaning center for american progress suggested \textcolor{red}{135 million people.}. also, it is unclear how many people with preexisting conditions would be at risk of losing their insurance entirely, or facing higher costs or having their conditions excluded from coverage. though bidens number is \textcolor{teal}{certainly within this range}, he would need to provide more detail to support such a definitive number. the hhs issue brief, published in january 2017, estimated that between \textcolor{red}{61 million and 133 million} americans have a preexisting condition.
 \paragraph{[\textbf{Verdict}]:}\textcolor{teal}{\textsf{True}} \\

\textbf{LLM reasoning without \name{}} \\
\textbf{[Reasoning]:}  The evidence suggests a \textcolor{red}{range of estimates for the number of people with pre-existing conditions}, with Biden's number being within this range but not universally accepted.

\textbf{[Prediction]:} \textcolor{red}{Conflicting} \\

\textbf{LLM reasoning  with \name{}} 

 \textbf{[Reasoning]:}  The claim is supported by multiple credible sources, including a former Vice President and health care consulting firms, with estimates \textcolor{teal}{ranging from 61 to 135 million} Americans having pre-existing conditions. \textcolor{teal}{Biden's statement of 100 million falls within this range.}

 \paragraph{[\textbf{Prediction}]:}\textcolor{teal}{\textsf{True}}
\end{tcolorbox}

\vspace{-1em}
\captionof{figure}{Example numerical claim demonstrating \textit{\analysisissue} issue and mitigation offered by \name{}}\label{tab:overanalysis_example1} 
\end{table}

\begin{table}[htb] 
\centering
\small
\begin{tcolorbox}[title=Mitigating \textit{\analysisissue} for Numerical FC]
\small
\medskip\noindent \paragraph{\textbf{Claim}:}\texttt{Egas Moniz, the inventor of the controversial lobotomy technique for treating mentally ill patients (particularly people with schizophrenia), was awarded a Nobel Prize in 1949 for his work.}

 \paragraph{[\textbf{Evidence}]:} antnio egas moniz (1874-1955) of portugal was an ambitious and multitalented person. ... \textcolor{red}{he had hoped and perhaps expected to receive the nobel prize for this work}, and \textcolor{red}{was disappointed when he realized he would not.} in 1935 at an international neurology conference he saw a ... loses something by this operation, some spontaneity, some sparkle, some flavor of the personality." in 1942 they published an influential book promoting the practice. in the united states the number of lobotomies performed per year went from 100 in 1946 to 5,000 in \textcolor{teal}{1949. that year moniz won the nobel prize in physiology/medicine for his contribution} .... reduced the perceived need for most lobotomies.

 \paragraph{[\textbf{Verdict}]:}\textcolor{teal}{\textsf{True}} \\

\textbf{LLM reasoning without \name{}} \\
\textbf{[Reasoning]:}  The text states that Moniz was \textcolor{red}{disappointed when he realized he would not receive the Nobel Prize} for his work on visualizing blood vessels in the brain, but later won the Nobel Prize in Physiology or Medicine in 1949. This suggests that there is a \textcolor{red}{discrepancy between the information provided.}

\textbf{[Prediction]:} \textcolor{red}{Conflicting} \\

\textbf{LLM reasoning  with \name{}} 

 \textbf{[Reasoning]:}  The text states that Moniz was disappointed when he realized he would not receive the Nobel Prize for his work on visualizing blood vessels in the brain, but \textcolor{teal}{later won the Nobel Prize in Physiology/Medicine in 1949} for his contribution to psychosurgery.

 \paragraph{[\textbf{Prediction}]:}\textcolor{teal}{\textsf{True}}
\end{tcolorbox}

\vspace{-1em}
\captionof{figure}{Example numerical claim demonstrating \textit{\analysisissue} issue and mitigation offered by \name{}}\label{tab:overanalysis_example2} 
\end{table}

\section{Prompts}
\label{appendix:prompts}

This section describes the system and user prompts employed during different phases of our experiments. Figures \ref{tab:system_prompt_llm_judge} and \ref{tab:overanalysis_1} show the system prompt and a sample user prompt used to determine whether verifier-based reasoning reduces \textit{\analysisissue} compared to its Top 1 Decoding counterpart. Figure \ref{fig:fact_checking_prompt} illustrates an example prompt designed to guide the fact-checking LLM in predicting a verdict and providing reasoning for a claim-evidence pair within a few-shot learning framework. Lastly, Figure \ref{tab:decomposition_prompt} presents the system prompt used to generate sub-questions by breaking down a claim, following the methodology detailed in the CLAIMDECOMP paper \cite{claimdecomp}.

\begin{table*}[htbp]
\centering
\small
\begin{tcolorbox}[title=System Prompt to Identify \analysisissue]

\paragraph{\textbf{Instruction:}} 
You are a fact-checking expert tasked with evaluating whether the reasoning generated using a verifier model mitigates a common issue in language model inference known as \textit{\analysisissue}.

You will be provided with:
\begin{itemize}
    \item A \textbf{claim}
    \item Its corresponding \textbf{evidence}
    \item The \textbf{verdict} for the claim
    \item A reasoning and prediction generated \textbf{without} a verifier model
    \item A reasoning and prediction generated \textbf{with} a verifier model
\end{itemize}

\textbf{Goal:}  
Determine whether the verifier-based reasoning successfully mitigates \analysisissue. \analysisissue is defined as:

\begin{itemize}
    \item Focusing on details in the evidence that are not pertinent to the claim’s veracity.
    \item Being overly strict with numerical or factual precision that could be interpreted more leniently.
    \item Verifying sub-questions individually but failing to aggregate them into a coherent, accurate final verdict.
\end{itemize}

\textbf{Verifier reasoning is considered to mitigate \analysisissue if:}
\begin{itemize}
    \item It focuses only on evidence that is relevant to the claim.
    \item It avoids unnecessary scrutiny of minor or irrelevant details.
    \item It improves label accuracy by reasoning holistically about the claim.
\end{itemize}

\textbf{Your task:}  
Return a binary label based on your assessment:

\begin{itemize}
    \item \texttt{1} — \analysisissue is mitigated in the verifier-based reasoning.
    \item \texttt{0} — \analysisissue is not mitigated (i.e., both approaches suffer equally or the verifier reasoning also overanalyzes).
\end{itemize}

\textbf{Output format:}
\begin{enumerate}
    \item \textbf{Label:} \texttt{<0 or 1>}
    \item \textbf{Explanation:} \texttt{<Brief justification referencing key differences in reasoning and use of evidence>}
\end{enumerate}

\end{tcolorbox}

\captionof{figure}{System prompt provided to the judge LLM to evaluate and identify \analysisissue between reasoning with and without verifier support.}
\label{tab:system_prompt_llm_judge}

\end{table*}

\begin{table*}[htbp] 
\centering
\small
\begin{tcolorbox}[title=An example user prompt to identify \analysisissue]
\medskip\noindent 

\textbf{\large \textbf{Example:}}

\vspace{0.3em}
[Claim]: Joe Biden says, “There's 100 million people that have preexisting conditions.”

\vspace{0.3em}
[Evidence]: a 2018 analysis by avalere, a health care consulting firm, estimated that 102 million americans have preexisting conditions. ... hhs issue brief, published in january 2017, estimated that between 61 million and 133 million americans have a preexisting condition.

\vspace{0.3em}
[Verdict]: True

\vspace{0.6em}
\textbf{LLM reasoning without verifier}

\vspace{0.3em}
[Reasoning]: The evidence suggests a range of estimates for the number of people with pre-existing conditions, with Biden's number being within this range but not universally accepted.

\vspace{0.3em}
[Prediction]: Conflicting

\vspace{0.6em}
\textbf{LLM reasoning with verifier}

\vspace{0.3em}
[Reasoning]: The claim is supported by multiple credible sources, including a former Vice President and health care consulting firms, with estimates ranging from 61 to 135 million Americans having pre-existing conditions. Biden's statement of 100 million falls within this range.

\vspace{0.3em}
[Prediction]: True

\vspace{0.6em}
\textbf{LLM judgement}

\vspace{0.3em}
[Label]: 1

\vspace{0.3em}
[Explanation]:  The reward model-based reasoning effectively mitigates overthinking by focusing on the relevant evidence that supports the claim, specifically highlighting that Biden's figure of 100 million falls within a credible range of estimates (61 to 135 million). In contrast, the reasoning without the reward model gets bogged down in the conflicting nature of the estimates without recognizing that Biden's number is still valid within that context. The reward model reasoning arrives at the correct label (TRUE) more reliably by synthesizing the evidence holistically rather than getting lost in the details of conflicting statements.

 \vspace{1.5em}

\textbf{\large \textbf{Input:}}

\vspace{0.3em}
[Claim]: Diaspora remittances “now stand at an all-time high of over KSh400 billion in 2021”.

\vspace{0.3em}
[Evidence]: Diaspora remittances were robust in 2021, with total remittances in 2021 reaching a record USD 3,718 million.

\vspace{0.3em}
[Verdict]: True

\vspace{0.6em}
\textbf{LLM reasoning without verifier}

\vspace{0.3em}
[Reasoning]: The evidence provides conflicting information about the exact amount of diaspora remittances in 2021, and some sources mention figures that are not directly comparable to KSh400 billion.

\vspace{0.3em}
[Prediction]: Conflicting

\vspace{0.6em}
\textbf{LLM reasoning with verifier}

\vspace{0.3em}
[Reasoning]: The evidence supports the claim that diaspora remittances reached an all-time high of over KSh400 billion in 2021. However, there are minor variations in the wording and units used across different sources.

\vspace{0.3em}
[Prediction]: True

\end{tcolorbox}

\vspace{-1em}
\captionof{figure}{An example user prompt to the judge LLM to identify \analysisissue between the reasoning without verifier and with verifier}\label{tab:overanalysis_1}
\end{table*}

\begin{table*}[htbp] 
\centering
\small
\begin{tcolorbox}[title=An example prompt to generate reasoning sequences for a claim - evidence pairs ]
\medskip\noindent 

\textbf{\large \textbf{System prompt:}}

\vspace{0.3em}

You are a fact-checking assistant tasked with evaluating claims based on provided evidence. Your task:
    \begin{enumerate}
        \item  Carefully analyze the important details from both the claim and the evidence provided. 
        \item Reason through the evidence step-by-step, synthesizing relevant information to assess the overall veracity of the claim. 
        \item Finally Classify the claim's veracity into one of the following categories:
        \begin{enumerate}
            \item True: The evidence fully supports the claim.
            \item False: The evidence directly contradicts the claim.
            \item Conflicting: The evidence contains contradictory elements or is inconclusive.
        \end{enumerate}
    \end{enumerate}   \
    \\
    **Response Format**:

    - [Label]: (True, False, or Conflicting)
    
    - [Justification]: The reasoning steps used that led to your classification of the claim.

 \vspace{1.5em}

\textbf{\large \textbf{User prompt:}}

\vspace{0.3em}
\textbf{Example 1}

\vspace{0.3em}

[Claim]: Argentina's Central Bank is considering putting Lionel Messi's image on the 1000 pesos bank note.

\vspace{0.3em}
[Evidence]: in fact, such was his influence in the argentina camp, the country's central bank are thinking about ... governing body are considering immortalising world cup winner lionel messi by putting his face on a 1000-peso note, according to reports.

\vspace{0.3em}
[Verdict]: False

\vspace{0.3em}
[Justification]: Is it true? The claim is false. Neither does the BCRA's website show Messi's picture on any banknote, nor has it made such an announcement.

\vspace{0.6em}
\textbf{Example 2}

[Claim]: Politico did not report that 20 million Ukrainians would die to win war.

\vspace{0.3em}
[Evidence]: in the ongoing russia-ukraine conflict, losses reported by different sources vary by orders of magnitude. sergei shoigu reported 5,937 russian soldier ... russian soldier killed, russian sources report only 0.3 losses. on the contrary, for every ukrainian soldier killed, russian sources report 4.3 ukrainian soldiers killed.

\vspace{0.3em}
[Verdict]: True

\vspace{0.3em}
[Justification]: Politico did not publish an article saying the deaths of 20 million Ukrainians were needed to win Russia 2019s .... The report cannot be seen when searching through Politico 2019s social media accounts.

\vspace{0.6em}

\textbf{Input prompt}

 Similar to the given examples, fact check the following claim using the evidence. Pay additional attention to numerical spans in claim and evidence and fact check by thinking step by step and output the label by performing entailment. Classify the entire claim strictly into one of the following categories: TRUE, FALSE or CONFLICTING.

\vspace{0.6em}
[Claim]: Diaspora remittances “now stand at an all-time high of over KSh400 billion in 2021”.

\vspace{0.3em}
[Evidence]: Diaspora remittances were robust in 2021, with total remittances in 2021 reaching a record USD 3,718 million.

\vspace{0.3em}
Give the reply in the following format:

\vspace{0.3em}

[Label]: (SUPPORTS, REFUTES or CONFLICTING).

[Justification]:
 
\end{tcolorbox}

\vspace{-1em}
\captionof{figure}{An example prompt to generate veracity prediction and corresponding reasoning sequence for a claim, evidence pair.}\label{fig:fact_checking_prompt} 
\end{table*}

\begin{table*}[htbp] 
\centering
\small
\begin{tcolorbox}[title= Prompt to decompose a claim ]
\medskip\noindent 

\textbf{\large \textbf{System prompt:}}

\vspace{0.3em}
You are tasked with generating a set of questions to break down and evaluate the veracity of a given fact-checking claim. Follow these strict guidelines when creating the subquestions:
    \begin{enumerate}
        \item  \textbf{Comprehensive Coverage:} The subquestions should comprehensively address all relevant aspects of the claim to enable a thorough fact-check.
        \item \textbf{Non-redundancy:} Ensure there is no overlap between the subquestions. Each question should be unique in meaning and not semantically repetitive.
        \item \textbf{Relevance:} Every question must be relevant in determining the veracity of the claim. Avoid irrelevant or tangential questions.
        \item \textbf{Clarity and Grammar:} Write questions that are clean, concise, and grammatically correct to maintain clarity.
        \item \textbf{Yes/No Format:} Each question must be strictly answerable with a simple "Yes" or "No" response.
    \end{enumerate}

\end{tcolorbox}

\vspace{-1em}
\captionof{figure}{ System prompt to generate sub-questions given a claim, adhering to the
standards defined in the CLAIMDECOMP paper.}\label{tab:decomposition_prompt} 
\end{table*}

\end{document}